\documentclass[10pt,twocolumn,letterpaper]{article}

\usepackage{cvpr}
\usepackage{times}
\usepackage{epsfig}
\usepackage{graphicx}
\usepackage{caption}
\usepackage{amsmath}
\usepackage{amssymb}
\DeclareMathOperator{\E}{\mathbb{E}}
\DeclareMathOperator*{\argmin}{argmin} 

\usepackage{multirow}
\usepackage{subfigure}
\usepackage{epstopdf}
\usepackage{epigraph}
\usepackage{etoolbox}
\makeatletter
\patchcmd{\epigraph}{\@epitext{#1}}{\itshape\@epitext{#1}}{}{}
\makeatother

\usepackage{array}
\newcommand{\calcfactor}[1]{%
  \dimexpr#1\textwidth-2\tabcolsep-1.5\arrayrulewidth\relax
}
\newcolumntype{P}[1]{p{\calcfactor{#1}}}

\cvprfinalcopy 

\def\cvprPaperID{1185} 

\ifcvprfinal\fi
\begin{document}

\newcommand{\todo}[1]{\textcolor[rgb]{1,0,0}{#1}}

\title{Learning to Sketch with Shortcut Cycle Consistency}

\author{Jifei Song$^1$ \quad Kaiyue Pang$^1$ \quad Yi-Zhe Song$^1$ \quad Tao Xiang$^1$ \quad Timothy M. Hospedales$^{1,2}$\\
$^1$SketchX, Queen Mary University of London \quad \quad 
$^2$The University of Edinburgh \\
{\tt\small \{j.song, kaiyue.pang, yizhe.song, t.xiang\}@qmul.ac.uk, 
t.hospedales@ed.ac.uk}
}

\twocolumn[{%
\renewcommand\twocolumn[1][]{#1}%
\maketitle
\thispagestyle{empty}
\begin{center}
    \centering
    \includegraphics[width=0.95\textwidth]{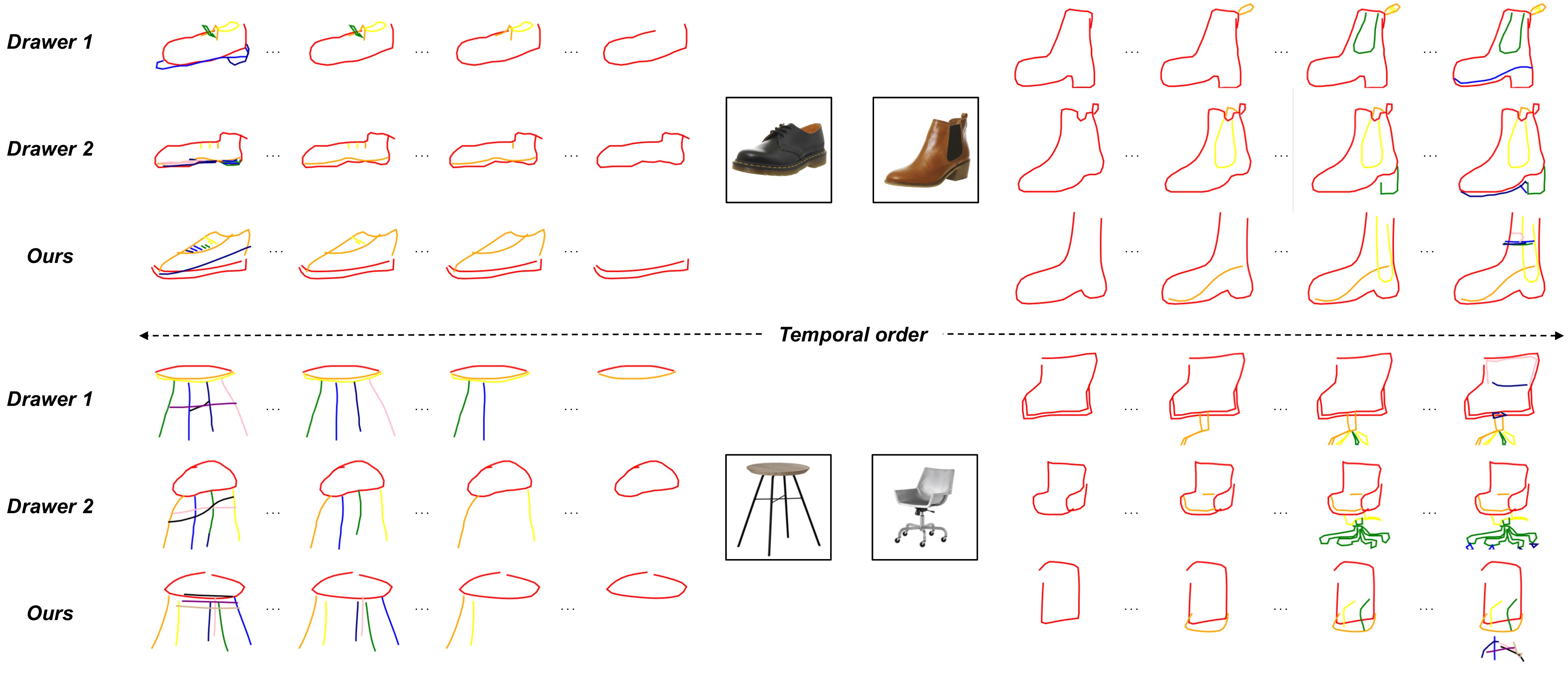}
    \captionof{figure}{Given one object photo, our model learns to sketch stroke by stroke, abstractly but semantically,  mimicking human visual interpretation of the object. Our synthesized sketches maintain a noticeable difference from human sketches rather than simple rote learning (\eg, shoelace for top left shoe, leg for bottom right chair). Photos presented here have never been seen by our model during training. Temporal strokes are rendered in different colors. Best viewed in color.}
     \label{fig:highlight}
\end{center}%
}]

\begin{abstract}
To see is to sketch -- free-hand sketching naturally builds ties between human and machine vision.   In this paper, we present a novel approach for translating an object photo to a sketch, mimicking the human sketching process. This is an extremely challenging task because the photo and sketch domains differ significantly. Furthermore,  human sketches exhibit various levels of sophistication and abstraction even when depicting the same object instance in a reference photo. This means that even if photo-sketch pairs are available, they only provide weak supervision signal to learn a translation model. Compared with existing supervised approaches that solve the problem of $D(E(photo))\rightarrow sketch$), \textcolor{black}{where $E(\cdot)$ and $D(\cdot)$ denote encoder and decoder respectively,}  we take advantage of the inverse problem (\eg, $D(E(sketch)\rightarrow photo$), and combine with  the unsupervised learning tasks of within-domain reconstruction, all within a  multi-task learning framework. Compared with existing  unsupervised approaches based on  cycle consistency (\ie, $D(E(D(E(photo))))\rightarrow photo$), we introduce a shortcut consistency enforced at the encoder bottleneck (\eg, $D(E(photo))\rightarrow photo$) to exploit the additional self-supervision. Both qualitative and quantitative results show that the proposed model is superior to a number of state-of-the-art alternatives. We also show that  the synthetic sketches can be used to train a better fine-grained  sketch-based image retrieval (FG-SBIR) model, effectively alleviating the problem of sketch data scarcity.
 \end{abstract}
 
\section{Introduction}

What do we see when our eyes perceive a grid of pixels from a real-world object? We can quickly answer this question by sketching a few line strokes.  Despite the fact that drawings like this may not exactly match the object as captured by a photo, they do tell us how we perceive and represent the visual world around us, that is, we as humans convey our perception of objects abstractly but semantically. In this context, it is natural to ask to what extent a machine can see. For decades, researchers in computer vision have dedicated themselves to answering this question, by injecting intelligence and supervision into the machine with the hope of seeing better. This is mostly done by formulating several specific constrained problems, such as classification, detection, identification, and segmentation.

In this paper, we take one step forward -- teaching a machine to generate a sketch from a photo just like humans do. This requires not only developing an abstract concept of a visual object instance, but also knowing what, where and when to sketch the next line stroke. Figure \ref{fig:highlight} shows  that the developed photo-to-sketch synthesizer takes a photo as input and mimics the human sketching process by sequentially drawing one stroke at a time. The resulting synthesized sketches provide an abstract and semantically meaningful depiction of the given object, just like human sketches do. 

\begin{figure}[t]
\centering
\includegraphics[width=0.45\textwidth]{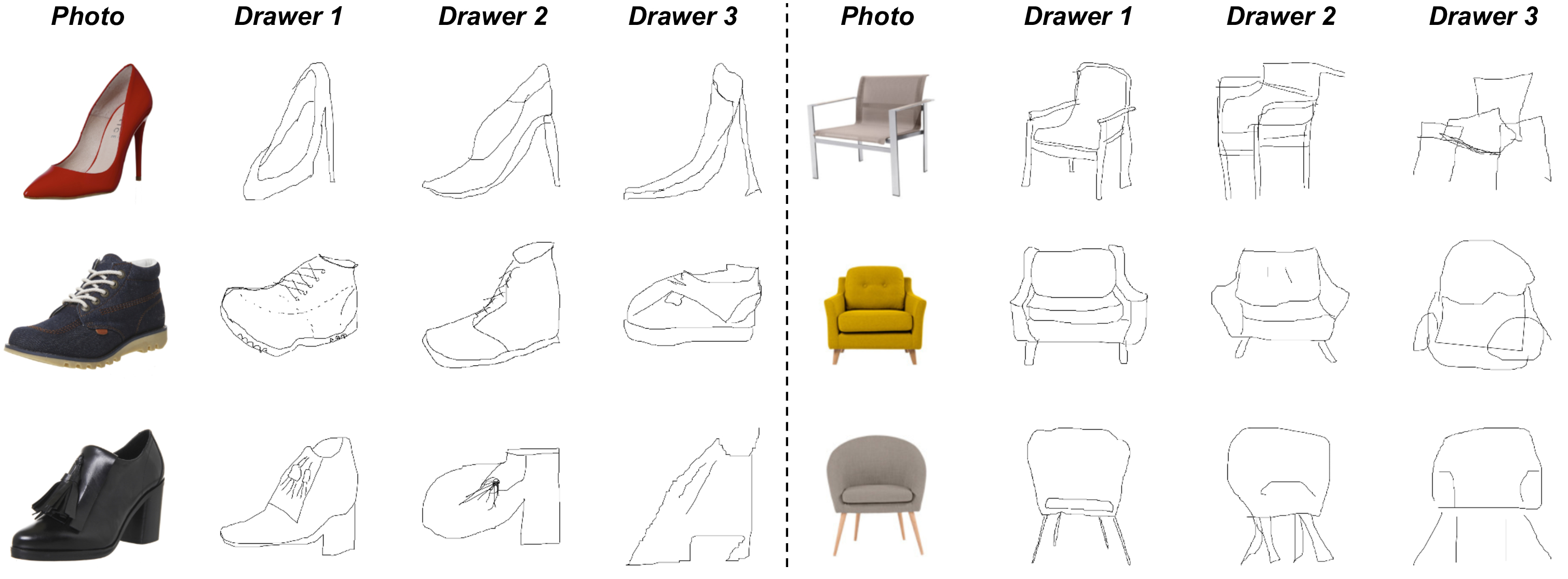}
\caption{Given a reference photo, sketches drawn by different people exhibit large variation in style and abstraction levels. Some of them are poor in depicting the object instances in the corresponding photos. }
\vspace{-0.5cm}
\label{fig:comparison}
\end{figure}

\begin{figure*}[tbp]
\centering
\includegraphics[width=0.85\textwidth]{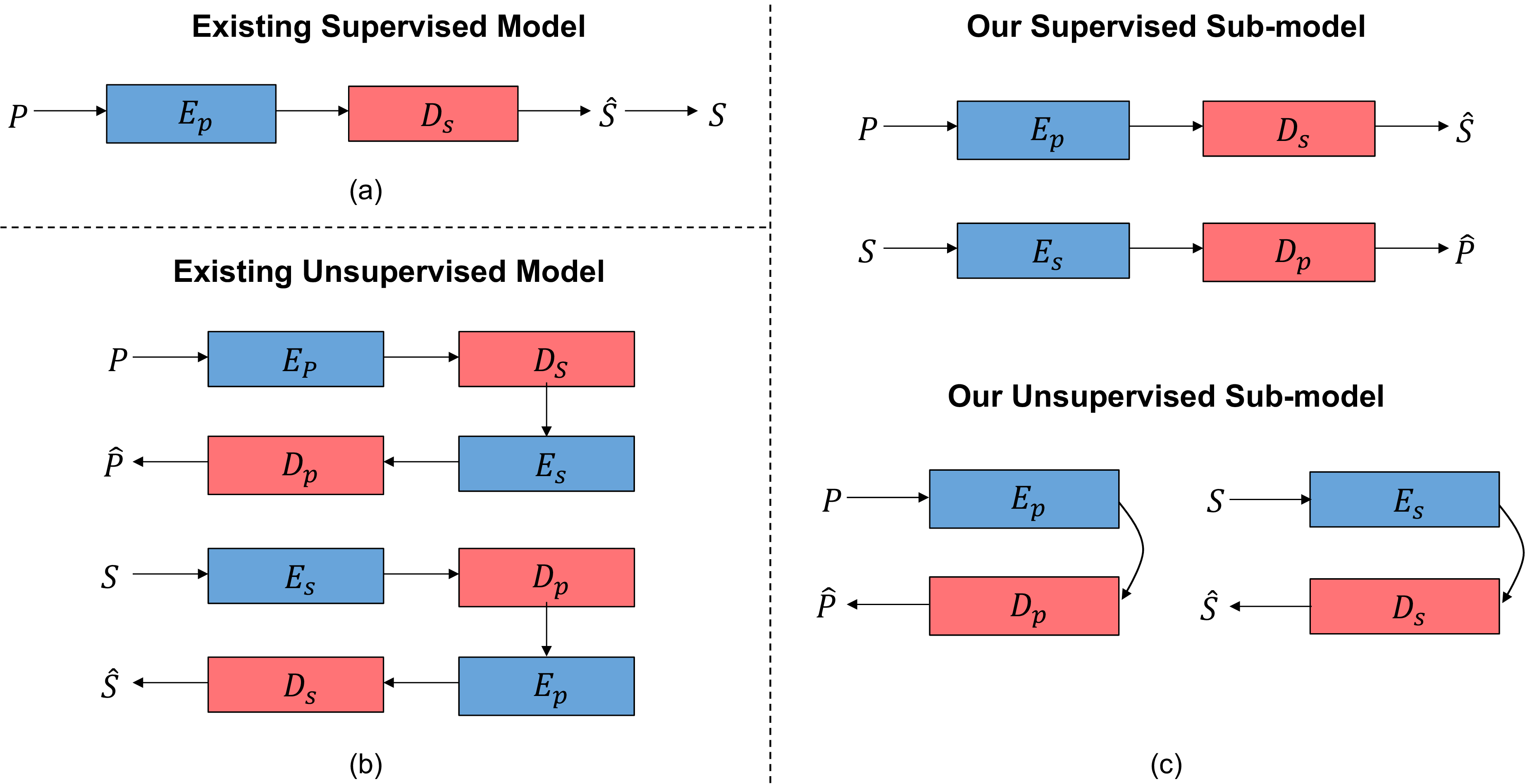}
\caption{(a) Existing supervised image-to-image translation framework, where mapping is one-way only. (b) Existing unsupervised image-to-image translation models enforce cycle consistency to address the highly under-constrained one-to-one mapping problem. (c) Our supervised-unsupervised hybrid model with dual/two-way supervised translation sub-models and two unsupervised sub-models with shortcut cycle consistency. This takes advantage of the noisy supervision signal offered by photo-sketch pairs, as well as learning from within-domain reconstruction.}
\label{fig:framework}
\end{figure*}

Photo-to-sketch synthesis can be considered as a cross-domain image-to-image translation problem. Thanks to the seminal work of \cite{sketchrnn,graves2013generating}, we are able to construct a  generative sequence model with recurrent neural network (RNN) acting as a neural sketcher. However, the synthesized sketches are not conditional on specific object photos. To address this problem, one can encode the photo via a convolutional neural network (CNN) and feed the code into the neural sketcher. Such a photo-to-sketch synthesizer essentially follows the traditional encoder-decoder architecture (see Figure \ref{fig:framework}(a)), and has been taken by most existing image-to-image translation models \cite{isola2016image, ledig2017photo}. Training such a model is done in a supervised manner requiring cross-domain image pairs: in our problem, these are photo-sketch pairs containing the same object instances. Compared to image-to-image translation, the key challenge for learning instance-level photo-to-sketch synthesis is that training pairs provide highly noisy supervision: Different sketches of the same photo have large style and abstraction differences between them (see Figure \ref{fig:comparison}). This makes our problem highly noisy and under-constrained.



In order to achieve photo-to-sketch synthesis with noisy photo-sketch pairs as supervision, we address the limitations of existing cross-domain image translation models by proposing a novel framework based on multi-task supervised and unsupervised hybrid learning (see Figure \ref{fig:framework}(c)). Taking an encoder-decoder architecture, our primary task is $D(E(photo)) \rightarrow sketch$) where a photo is first encoded by $E$ and then decoded into a sketch by $D$. To help learn a better encoder and decoder, we introduce the inverse problem ($D(E(sketch))\rightarrow photo$) so that the supervised model learning can be done in both directions. Importantly, we also introduce two unsupervised learning tasks for within-domain reconstruction, \ie,  $D(E(photo))\rightarrow photo$ and  $D(E(sketch))\rightarrow sketch$. This hybrid learning framework differs significantly from existing approaches in that: (1) It combines supervised and unsupervised learning in a multi-task learning framework in order to make the best use of the noisy supervision signal. In particular, by sharing the encoder and decoder in various tasks, a more robust and effective encoder and decoder for the main photo-to-sketch synthesis task can be obtained. (2)  Different from the existing unsupervised models based on cycle consistency (Figure \ref{fig:framework}(b)), our unsupervised learning tasks exploit the notion of shortcut cycle consistency: instead of passing through a different domain to get back to the input domain for reconstruction, our model takes a shortcut and completes a reconstruction within each domain. This is particularly effective given the large domain gap between photo and sketch. 

Figure \ref{fig:highlight} shows that our model successfully translates photo to sketches stroke by stroke, demonstrating that the model has acquired an abstract and semantic understanding of visual objects. We compare against a number of state-of-the-art cross-domain image translation models, and show that superior performance is obtained by our model due to the proposed novel supervised and unsupervised hybrid learning framework with the  shortcut cycle consistency. We also quantitatively validate the usefulness of the  synthesized  sketches for training a better fine-grained sketch-based image retrieval  (FG-SBIR) model.

Our contribution is summarized as follows:
(1) To our best knowledge, for the first time, the photo-to-sketch synthesis problem is addressed using a \textit{learned} deep model, which enables stroke-level cross-domain visual understanding from a reference photo.
(2) We identify the noisy supervision problem caused by subjective and varied human drawing styles, and propose a novel solution with hybrid supervised-unsupervised multi-task learning. The unsupervised learning is accomplished more effectively via a shortcut cycle consistency constraint. 
(3) We exploit the synthesized sketches as an alternative to  expensive photo-sketch pair annotation for training a FG-SBIR model. Promising results are obtained by using the synthesized photo-sketch pairs to augment manually collected pairs. 

\section{Related Work}

\noindent \textbf{Image-to-Image Translation} Recent advances on generative modeling make realistic image generation possible.  Image generation can be conditional on class labels \cite{mirza2014conditional}, attributes \cite{yan2016attribute2image, lample2017fader}, text \cite{reed2016generative, zhang2017stackgan} and images \cite{isola2016image, ledig2017photo, hu2017}. For image-to-image generation/translation, if paired data (input and output image) are available, most recent approaches adopt a conditional generative adversarial network (GAN), from which a joint distribution is readily manifested and can be matched to the empirical joint distribution provided by the paired data. However for many tasks, paired data are often difficult to acquire for supervised learning; unsupervised learning methods thus started to get popular recently. BiGAN \cite{donahue2017adversarial} and ALI \cite{dumoulin2017adversarially} are models that jointly learn a generation network and inference network via adversarial learning. Other models including  DiscoGAN \cite{kim2017learning}, CycleGAN \cite{zhu2017unpaired} and DualGAN \cite{yi2017dualgan} adopted two generators to model the bidirectional mapping between domains with adversarially trained discriminators to identify each. Cycle consistency is further added as a way to transitively regularize structured data, which greatly alleviates \textit{non-identifiability} issues \cite{li2017towards}. Additional weight-sharing constraints are also explored in CoGAN \cite{liu2016coupled} and UNIT \cite{liu2017unsupervised} to build a bond between domain marginal distributions. Note that most previous works rely on the assumption of level of pixel-to-pixel correspondence to a certain extent,  which clearly does not hold for our sketch-to-photo translation problem. In our problem, pairwise supervision is available but the supervision signal is noisy and weak, challenging the existing supervised learning based methods. Nevertheless this supervision is too useful to ignore by adopting an entirely unsupervised learning approach. Therefore we propose a novel hybrid model to have the best of both worlds.  

\noindent \textbf{Recurrent Vector Image Generation}\quad Most recent image generation and understanding work generate images in a continuous pixel space via convolutional neural networks (CNNs) \cite{isola2016image, zhu2016generative, zhang2017stackgan, lample2017fader}. There has been relatively few studies on vector image generation. Vector representation is perfectly suited for sketches because both spatial and temporal visual cues are encoded during the sketching process. The seminal work of Graves \emph{et al.} \cite{graves2013generating} adopted recurrent neural networks (RNNs) to generate vector handwritten digits by using mixture density networks for continuous data points approximation. Similar models were developed for vectorized Kanji characters \cite{zhang2017drawing, kanjiweb} and free-hand human sketches \cite{sketchrnn}, both conditionally and unconditionally by modeling them as a sequence of pen stroke actions. Very recently, \cite{chen2017sketcH} proposed to build ties between raster and vector sketch images through a CNN-RNN paradigm. In this work, sketches are stored as vector images and a RNN decoder is employed to generate sketches from a CNN encoder embedding, resulting in clean and sharp line strokes, which has shown better sketch generation performance compared to \cite{sketchrnn}. 

\noindent \textbf{Vector Sketch Datasets}\quad One main factor that hampers research on generating vector sketch images is the lack of publicly available large-scale datasets. The TU-Berlin \cite{eitz2012humans} and Sketchy \cite{sangkloy2016sketchy} datasets  provide 20k and $\sim$70k vector sketches from multiple categories respectively. They are designed for sketch recognition and FG-SBIR respectively. But they are not quite big enough for learning a sketch generation model.  The lack of data problem is partially solved  in \cite{sketchrnn}, which contributes a dataset of 50 million vector drawings covering hundreds of categories obtained from the QuickDraw AI Experiment \cite{quickdraw}. Nevertheless, these category-level symbolic and conceptual vector drawings were each sketched within 20 seconds, so they often do not possess sufficient fine-grained detail for distinguishing object instances belonging to the same category.  To our knowledge, the largest fine-grained paired sketch-photo dataset to date is the QMUL-Shoe-Chair-V2 dataset \cite{sketchx}, which contains over 8000 photo-sketch pairs from two categories. In this work we focus on these two categories and use the QuickDraw shoe and chair sketches \cite{sketchrnn} for pretraining, and QMUL-Shoe-Chair-V2 for model fine-tuning.

\noindent  \textbf{Learning Discriminative Models with Synthetic Data}\quad A number of recent studies use data synthesized using deep generative models for training discriminative models, therefore circumventing the need for large-scale manual data collection and annotation. These discriminative models have been applied to various tasks including gaze estimation \cite{shrivastava2017learning}, hand pose estimation \cite{tompson2014real, supancic2015depth} and human pose estimation \cite{park2015articulated}. The  most related work is \cite{yu2017semantic}, which controls the  variations in the synthesized images using a \textit{learned} deep model rather than heuristic rendering. Most existing works use synthesized photo images, whilst in this work we aim to use synthesized sketches to learn a  discriminative model.

\noindent \textbf{Fine-grained Sketch-based Image Retrieval}\quad  One such discriminative models is a fine-grained sketch-based image retrieval (FG-SBIR) model. FG-SBIR addresses the problem of finding a specific photo containing the same instance as an input sketch. The relevant research field has flourished recently \cite{li2014fine, yu2016sketch, song2016sketch, sangkloy2016sketchy, li2017synergistic, pang2017fgsbir, xu2017cross, song2017sketch} due to its huge potential for commercial applications. One primary challenge is how to train a  model with limited sketch-photo pairs, because collecting free-hand sketch-photo pairs is very  expensive in practice. Previous work \cite{yu2017sketch} resorts to heuristic stroke augmentation and removal techniques to enhance the training data. In this work, for the first time, we attempt to generate synthetic sketch drawings from a learned deep model to boost FG-SBIR performance.

\section{Methodology}

\subsection{Overview}
\label{sub:overview}
We aim to learn a mapping function between the photo domain $X$ and sketch domain $Y$, where we denote the  empirical data distribution as $x \sim p_{data}(x)$ and $y \sim p_{data}(y)$ and represent each vector sketch segment as $(s_{x_{i}}, s_{y_{i}})$, a two-dimensional offset vector. Our model includes four mapping functions, learned using four subnets namely a photo encoder, a sketch encoder, a photo decoder, a sketch decoder. They are denoted as $E_{p}$, $E_{s}$, $D_{p}$ and $D_{s}$ respectively. 

\noindent\textbf{Sub-Models}\quad As illustrated by Figure \ref{fig:framework}(c), our model consists of four sub-models, each comprising an encoder subnet and a decoder subnet.  (1) A supervised sub-model that translates a photo to a sketch; (2) a supervised sub-model that  maps a sketch back to the photo domain; (3) an unsupervised sub-model to reconstruct photo and (4) an unsupervised sub-model to reconstruct sketch. This means that our learning objective consists of two types of losses (to be detailed later): supervised translation loss for matching cross-domain and shortcut cycle consistency loss for traversing within domain. 

\begin{figure*}[t]
\centering
\includegraphics[width=0.85\textwidth]{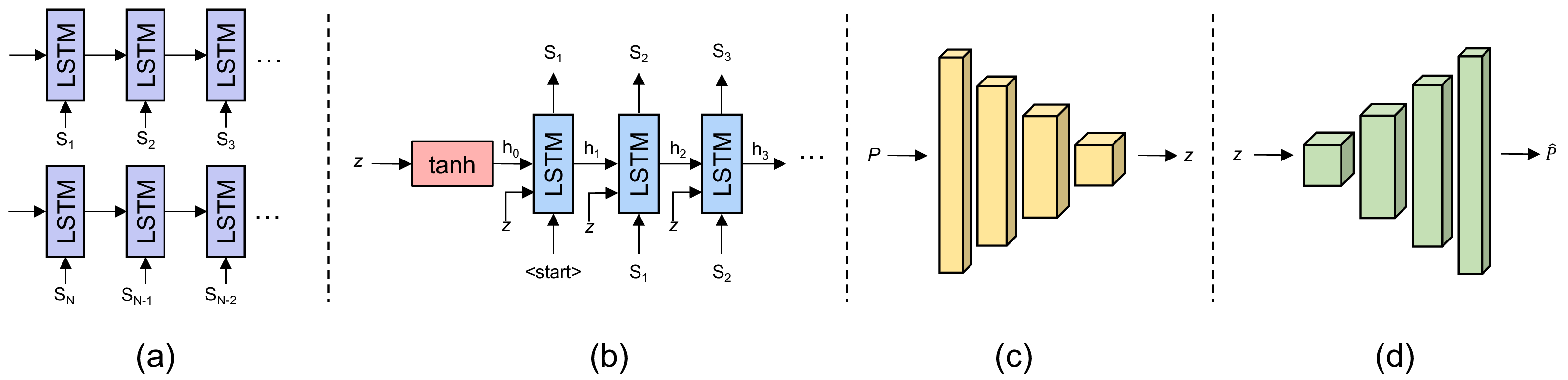}
\caption{(a) bidirectional LSTM encoder $E_{s}$. (b) conditional LSTM decoder $D_{s}$. (c) generative CNN encoder $E_{p}$. (d) conditional CNN decoder $D_{p}$.}
\label{fig:architect}
\end{figure*}

\noindent\textbf{Variational Encoders}\quad The two encoders $E_{p}$ and $E_{s}$ are CNN and RNN respectively (see Figures~\ref{fig:architect}(a) and (c)). In particular, $E_{s}$ is a bidirectional LSTM. They take in either a photo or sketch as input and output a latent vector. They are variational because the latent vector is then projected into two vectors $\mu$ and $\sigma$ with one fully connected (FC) layer. From the FC layer we construct our final embedding layer (bottleneck layer in each sub-model) by fusing it with a random vector, $\mathcal{N}(0, I)$, sampled from IID Gaussian distribution. To enable efficient posterior sampling, the re-parameterization trick is used as in \cite{kingma2013auto}:

\begin{equation}
\label{equ:z_vec}
z = \mu + \sigma \odot \mathcal{N}(0, I)
\end{equation}

\noindent\textbf{Sketch Decoder}\quad We build an LSTM-based sequence model as in \cite{sketchrnn} to sample output sketches segment by segment conditioned on the latent vector $z$ (see Figure~\ref{fig:architect}(b)). This is done by predicting each sketch segment offset $p(\Delta s_{x_{i}},\Delta s_{y_{i}})$ using a Gaussian mixture model and modeling pen state $q_{i}$ for each time step as a categorical distribution. We refer the reader to \cite{sketchrnn} for more details. To train the LSTM decoder, the reconstruction loss is formulated as:
\begin{equation}
\label{equ:rnn_loss}
 \begin{aligned}
&\mathcal{L}_{rnn}(S, \hat{S}) = \E_{x\sim S, y\sim \hat{S}} \\ &\Big[-\frac{1}{N_{max}}\Big(\sum_{i=1}^{N_s}{{\log}{(p({\Delta}s_{x_{i}}, {\Delta}s_{y_{i}} \rvert x, y))}} \\
&- \sum_{i=1}^{N_{max}}{\sum_{k=1}^3{p_{k,i}\log(q_{k,i}\rvert x,y)}} \Big) \Big]
\end{aligned}
\end{equation}
where $N_{max}$ represents the maximum number of segments in one sketch in the training set, and $N_s$ denotes the actual length of segments for one particular sketch, thus $N_s$ is usually smaller than $N_{max}$. Index $i$ and $k$ indicate the time step and one of three pen states, respectively. \textcolor{black}{With the supervision of the reconstruction loss, the sketch decoder is able to predict the next stroke segment based on the strokes of previous time stamps.}

\noindent\textbf{Photo Decoder}
We use a CNN-based deconvolutional-upsampling block, as is commonly adopted by various generative tasks, where an ${l_{2}}$ loss
 \begin{equation}
\label{equ:l2_loss}
\begin{aligned}
\mathcal{L}_{\rightarrow p}(P, \hat{P}) = \E_{x\sim P, y\sim \hat{P}}[||x-y||_{2}]
\end{aligned}
\end{equation}
 is used to measure the difference, which often leads to a blurry effect, known as the \textit{regression to mean} problem \cite{mathieu2016deep}. An obvious solution is to add adversarial loss \cite{goodfellow2014generative} for obtaining shaper photo visual effect. This was however not adopted because: (a) We did not observe improved  photo-to-sketch synthesis, and even slightly worse due to the mode collapse issue, commonly observed with generative adversarial training \cite{salimans2016improved}. (b) Synthesizing photos is not the main goal of the model; it is used as an auxiliary task to help the main photo-to-sketch synthesis task.  

\subsection{Shortcut Cycle Consistency}

We might expect that learning a one-way mapping from photo to sketch should suffice, as paired examples exist for providing a supervision signal. However, as discussed, photo-sketch pairs provide a weak and noisy supervision signal, so such a one-way mapping function cannot be learned effectively. Our solution is to introduce two-way mapping using supervised learning and unsupervised reconstruction tasks. Since the four encoder and decoders are shared by these supervised and unsupervised tasks, they benefit from multi-task learning. 

For the under-constrained mapping in the unsupervised self-reconstruction tasks, cycle consistency \cite{he2016dual, zhu2017unpaired} is developed to alleviate the \textit{non-identifiable} \cite{li2017towards} problem by reducing the space of possible mappings. This is achieved from the intuition that for each source image, the translation should be cycle consistent as to bring back to itself from the translated target domain. Taking photo to sketch translation for example, we have $x \rightarrow E_{p}(x) \rightarrow D_{s}(E_{p}(x)) \rightarrow E_{s}(D_{s}(E_{p}(x)) \rightarrow D_{p}(E_{s}(D_{s}(E_{p}(x)))$. However, since we do have noisy but paired data to provide weak supervision, the approximate posterior can actually be learned within each domain from the encoder's embedding. This is achieved by enforcing a variational bound and this is exactly where the shortcut can happen in the new cycle consistency proposed in this work.

Specifically, to form a photo to photo cycle now requires only traverse within domain, \ie, $x \rightarrow E_{p}(x)\rightarrow D_{p}(E_{p}(x))$, which we term as shortcut cycle consistency. We  find that apart from resulting in faster convergence in our supervised-unsupervised hybrid framework, our unsupervised sub-models with the shortcut cycle consistency can produce much better photo-to-sketch synthesis compared with the model learned with conventional cycle consistency. We postulate that given the large domain gap between photo and sketch, doing a long walk across domains potentially makes it harder to establish cross-domain correspondence. Formally, to enforce the shortcut cycle consistency, we minimize the following loss:

 \begin{equation}
\label{equ:short_loss}
\begin{aligned}
\mathcal{L}_{shortcut}(X,Y) &=   \mathcal{L}_{\rightarrow s}(Y, D_{s}(E_{s}(Y))) \\&+ \mathcal{L}_{\rightarrow p}(X,D_{p}(E_{p}(X))) 
\end{aligned}
\end{equation}

\textcolor{black}{Note that although our shortcut consistency loss is formulated as a VAE type reconstruction loss, it serves a very different purpose here: to enforce consistency of the shared encoder and decoder for the cross-domain and cross-modality synthesis tasks.}

\subsection{Full Learning Objective}

The four sub-models are learned jointly. Therefore, in additional to the unsupervised loss above, there are thus two supervised translation losses:
 \begin{equation}
\label{equ:supervised_loss}
\begin{aligned}
\mathcal{L}_{supervised}(X,Y) &=  \mathcal{L}_{\rightarrow s}(Y,D_{s}(E_{p}(X))) \\ &+ \mathcal{L}_{\rightarrow p}(X, D_{p}(E_{s}(Y)))
\end{aligned}
\end{equation}

Furthermore, to enable efficient posterior sampling, we add KL losses for the bottleneck layer embedding space distributions to force the four sub-models to use a similar distribution to feed to their decoders. For simplicity, we combine them into one term:
 
\begin{equation}
\begin{aligned}
\mathcal{L}_{KL}= & \E_{x\sim X, y\sim Y, \hat{x}\sim \hat{X}, \hat{y}\sim \hat{Y}}\\ &[-\frac{1}{2}(1 + \sigma^2 - \exp(\sigma))|x, y, \hat{x}, \hat{y}]
\end{aligned}
\label{equ:kl_loss}
\end{equation}

Our full objective thus becomes:

\begin{equation}
\label{equ:full_loss}
\begin{aligned}
\mathcal{L}_{full}&(X,Y) = {L}_{supervised}(X,Y) \\ &+ \lambda_{shortcut} \mathcal{L}_{shortcut}(X,Y) + \lambda_{KL}\mathcal{L}_{KL}
\end{aligned}
\end{equation}
where $\lambda_{shortcut}, \lambda_{KL}$ controls the relative importance of each loss. With the full loss, we aim to optimize:

 \begin{equation}
\label{equ:final_optimize}
\begin{aligned}
\argmin_{E_{p}, E_{s}, D_{p}, D_{s}}L_{full}(X, Y)
\end{aligned}
\end{equation}

\section{Experiments}

\subsection{Datasets and Settings}

\noindent\textbf{Dataset Splits and Preprocessing} \quad We use the publicly available QMUL-Shoe-Chair-V2 \cite{sketchx} dataset, the largest stroke-level paired sketch-photo dataset to date, to train and evaluate our deep photo-to-sketch synthesis model. There are 6,648 sketches and 2,000 photos for the shoe category, where we use 5,982 and 1,800 of which respectively for training  and the rest for testing. For chairs, we split the dataset as following strategy: 300/100 photos, 1275/725 sketches for training/testing respectively. It is guaranteed that each photo is paired with at least one human sketch. We scale and center the photos to $224 \times224$ pixels and pre-process original sketches via stroke removal and spatial sampling to reduce to number of segments to the level  suitable for LSTM-based modeling.

\noindent\textbf{Pretraining on QuickDraw Dataset} \quad Due to the limited number of sketch-photo pairs in QMUL-Shoe-Chair-V2, we pretrain our model with 70,000 shoe and 70,000 chair training sketches from the QuickDraw dataset \cite{sketchrnn}. Despite the fact that only abstract iconic vector sketches exist with no associated photos, we form our pretrained photos by transforming sketches to raster pixel images. \textcolor{black}{In this way, 70,000 vector-raster sketch pairs can be formed for model pretraining.}

\begin{figure*}[tbp]
\centering
\includegraphics[width=0.85\textwidth]{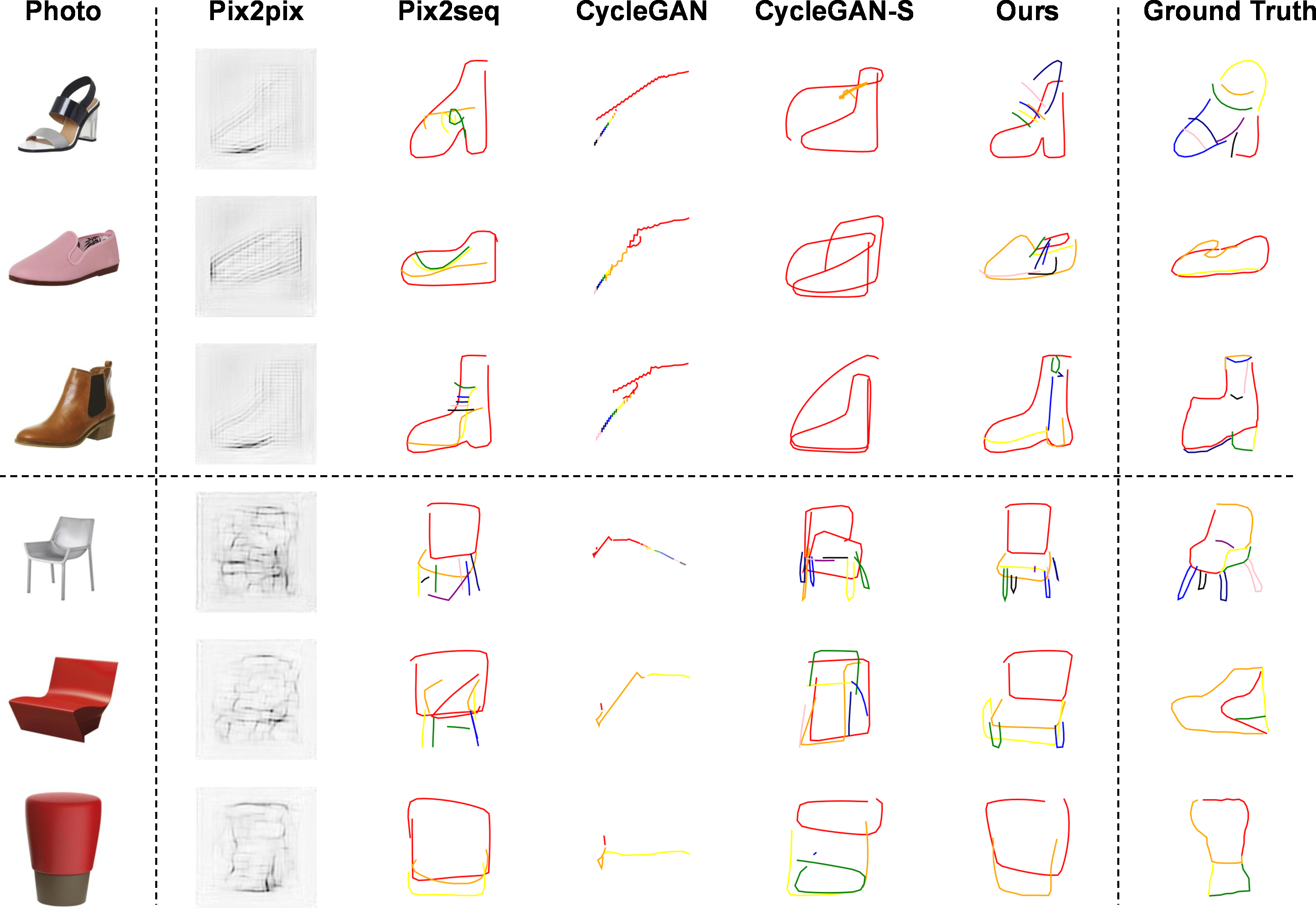}
\caption{Photo-to-sketch synthesis on the QMUL-Shoe-Chair-V2 test splits. From left to right: input photo, Pix2pix \cite{isola2016image}, Pix2seq \cite{chen2017sketcH}, CycleGAN \cite{zhu2017unpaired}, CycleGAN with supervised translation loss, ours and ground truth human sketch. Temporal strokes are rendered in different colors. Best viewed in color.}
\label{fig:results}
\vspace{-0.2em}
\end{figure*}

\noindent\textbf{Implementation Details}\quad Our CNN-based encoder and decoder, $E_p$ and $D_p$ consist of five stride-2 convolutions, two fully connected layers and five fractionally-strided convolutions with stride $1/2$, similar to \cite{isola2016image} but without skip connections. We use instance normalization instead of batch normalization as in \cite{ulyanov2016instance}. We adopt bidirectional and unidirectional LSTM for our RNN encoder $E_s$ and decoder $D_s$ respectively, while keeping other learning strategies the same as \cite{sketchrnn}. We implement our model end-to-end on Tensorflow \cite{abadi2016tensorflow} with a single Titan X GPU. We set the importance weights $\lambda_{shortcut} = 1$ and $\lambda_{KL} = 0.01$ during training and find this simple strategy works well. Both pretraining and fine-tuning stages are trained for a fixed 200,000 iterations with a batch size of 100. The model is trained end to end using the Adam optimizer \cite{kingma2014adam} with the parameters $\beta_{1} = 0.5, \beta_{2} = 0.9, \epsilon = 10^{-8}$. A fixed learning rate of 0.0001 is adopted for experiments.

\subsection{Evaluation Metric}
\label{sub: metric}
Evaluating the quality of synthesized images is still  an open problem. Traditional maximum likelihood approaches (\eg, kernel density estimation) fail to offer a true reflection of the synthesis quality, as validated in \cite{theis2016note}. Consequently,  most recent studies either run human perceptual studies by crowd-sourcing or explore computational metrics attempting to predict human perceptual similarity judgments \cite{odena2017conditional}. Our measures fall into the latter by discriminatively answering two questions: (i) How recognizable can the synthesized sketch be when evaluated with a recognition model trained on human sketch data? (ii) How realistic and diverse are the synthesized sketches, so that they can be used as queries to retrieve photos using a FG-SBIR model trained on  photo-human sketch pairs? A good score under these metrics requires synthesized sketches to be both realistic and instance-level identifiable. The metric thus shares the same intuition behind the ``inception score" \cite{salimans2016improved}. More specifically, the two metrics are: (1) \textbf{Recognition-Accuracy}: We feed the synthesized sketches into the sketch-a-net \cite{yu2017sketch} model, which is trained to recognize 250 real-world sketch categories with super-human performance. The assumption  is that if a  synthesized sketch can be recognized correctly as the same category as the corresponding photo, we can conclude with some confidence that it is category-level realistic. (2)
\textbf{FG-SBIR Retrieval-Accuracy}: Since our data are from the same category (either shoe or chair), the recognition-score could still be high if the model  learns to one specific object instance regardless of the input photo instances (\ie, the typical symptom of mode collapsing \cite{salimans2016improved}), or if the synthesized sketches are diverse but hardly resemble the object instances in the corresponding photos. To overcome this problem, the FG-SBIR accuracy  is  introduced as a harder metric. We retrain the model of \cite{yu2016sketch} on the QMUL-Shoe-Chair-V2 training split \cite{sketchx} and used the synthesized sketches to retrieve photos on the test split.

\subsection{Competitors}

For fair comparison, we implement all the competitors under the same architecture and training strategies as our model. 
\textbf{Pix2pix} \cite{isola2016image}: We compare with replacing vector sketch images with raster sketch images, where translation happens within the pixel space. We tried different state-of-the-art cross-domain translation models \cite{isola2016image, guccluturk2016convolutional, sangkloy2016scribbler}, but did not find much difference between them. We thus only report the results of the model in \cite{isola2016image} as a representative one.  \textbf{Pix2seq} \cite{chen2017sketcH}: This corresponds to the ablated version of our full model: a one-way photo-to-sketch supervised translation mode with vector sketch as output.  This is similar to \cite{chen2017sketcH}, which was originally designed  for better sketch reconstruction, now re-designed and re-purposed for the photo-to-sketch translation task. \textbf{CycleGAN} \cite{zhu2017unpaired}: This is proposed to specifically target image-to-image translation with the absence of paired training examples. Cycle consistency is enforced to alleviate the highly under-constrained setting of the problem. \textbf{CycleGAN-Supervised (CycleGAN-S)}: Additional supervised learning modules (two discriminators for adversarial training) are added on top of CycleGAN to give a level playing field. This can be considered as an alternative supervised-unsupervised hybrid model.

\vspace{0.3em}
\subsection{Qualitative Results}
\vspace{0.3em}
As illustrated in Figure \ref{fig:results}, all four competitors fail to generate high quality sketches that match with the corresponding photo. Our model, in contrast, is able to sketch object abstractly but semantically. Interestingly, our model produces some sketches with certain level of fine-grained details, which is extremely challenging given the highly noisy  supervision signals as shown in Figure \ref{fig:comparison}. In some cases, \eg, the third row shoe example, \textcolor{black}{the shape and the details of the synthesized sketch are more consistent to the reference photo, than those of human sketch.}


The competitors suffer from various problems. We observe complete model collapse when using CycleGAN under unsupervised setting, which suggests that CycleGAN may only works with unpaired training examples under a strong cross-domain pixel-level alignment assumption. After injecting supervision into CycleGAN (CycleGAN-S), the synthesized results get better but still suffers from regular noisy stroke generation, \ie, it seems that a random meaningless stroke is always sketched on a shoe. In contrast,  our model with shortcut cycle consistency does not suffer from such issue. This is because our model takes a shortcut from the bottleneck, which eases the burden on optimization and enhances the representation power of the encoder. We also witness some success using the Pix2seq model -- the sketch looks adequate on its own, but when compared with the corresponding photo, it does not bear much resemblance, often containing some wrong fine-grained details, \eg, ankle strap of the first-row shoe. This supports our hypothesis that one-way image-to-image translation  is not enough to deal with the highly-noisy paired training data. Finally, the worst results are obtained by the Pix2pix model which is the only model that treats sketch as a raster pixel image. The synthesized sketches are blurry and lack sharp and clean edges. This is likely caused by the fact that the model pays too much attention to handling the empty background which is also part of data to model with the raster image format. 

\subsection{Quantitative Results}
We compare  the performance of different models evaluated  using the two metrics (Sec.~\ref{sub: metric}) in Table \ref{tab:recog}. The following observations can be made: (i) Under the recognition metric, our model beats all the competitors. Interestingly it also beats human, showing our superior category-level generative realism. (ii) Under the retrieval metric, our model still outperforms all competitors on both datasets. However, this time, the gap to the human sketches' performance is big. This suggests that when humans draw a sketch of a specific object given a reference photo, attention is paid mainly to fine-grained details for distinguishing different instances, rather than the category-level realism. Nevertheless, compared to the chance level (0.5\%  acc.@1 for ShoeV2 and 1\% for ChairV2), our model's performance suggests the synthesized sketches do capture some instance-identifiable details.   (iii) The strongest competitor on ShoeV2  is Pix2seq \cite{chen2017sketcH}. However, its place is taken by CycleGAN-S on ChairV2. This is expected: the ChairV2 dataset is much smaller than ShoeV2, posing difficulties for a pure supervised-learning based approach. The unsupervised  CycleGAN yields poor performance all the time due to model collapse, but its supervised learning boosted version CycleGAN-S  fares quite well on the small ChairV2 dataset. This further validates our claims that a hybrid model is required and our shortcut consistency is more effective than the full cycle consistency. 




\begin{table}[t]
\small
\centering
\resizebox{\columnwidth}{!}{
\begin{tabular}{P{.18}|cP{.1}|cP{.1}cP{.1}cP{.1}}
\hline
 & \multicolumn{2}{c|}{Recognition} & \multicolumn{2}{c}{Retrieval}   \\
\hline
ShoeV2 & acc.@1 & acc.@10 & acc.@1 & acc.@10 \\
\hline
Human sketch \cite{sketchx} &36.50\% &70.00\%&\textcolor{black}{30.33\%} &\textcolor{black}{76.28\%} \\ \hline
Pix2pix \cite{isola2016image} & 0.00\% & 0.00\% & 0.50\% & 7.50\% \\
Pix2seq \cite{chen2017sketcH} &  \textcolor{blue}{51.50\%} & \textcolor{blue}{86.00\%} & \textcolor{blue}{4.50\%} & \textcolor{blue}{26.00\%} \\
CycleGAN \cite{zhu2017unpaired} & 0.00\% & 0.00\% & 0.50\% & 4.00\% \\
CycleGAN-S & 18.00\% & 51.50\% & 2.00\% & 18.00\% \\
Our full model & \textcolor{red}{53.50\%} & \textcolor{red}{90.00\%} & \textcolor{red}{6.00\%} & \textcolor{red}{28.50\%} \\
\hline
\hline
ChairV2 & acc.@1 & acc.@10 & acc.@1 & acc.@10 \\
\hline
Human sketch \cite{sketchx} &10.00\% &35.00\% &\textcolor{black}{47.68\%} &\textcolor{black}{89.47\%} \\ \hline
Pix2pix \cite{isola2016image} & 0.00\% & 0.00\% & 2.00\% & 16.00\% \\
Pix2seq \cite{chen2017sketcH} & 5.00\% &  \textcolor{blue}{51.00\%} & 3.00\% & 31.00\% \\
CycleGAN \cite{zhu2017unpaired} & 0.00\% & 8.00\%  & 1.00\% & 7.00\%\\
CycleGAN-S & \textcolor{blue}{12.00\%} & \textcolor{red}{55.00\%} & \textcolor{blue}{6.00\%} & \textcolor{blue}{33.00\%} \\
Our full model & \textcolor{red}{13.00\%} & \textcolor{red}{55.00\%} & \textcolor{red}{8.00\%} &\textcolor{red}{36.00\%} \\
\hline
\end{tabular}}
\protect\caption{Recognition and retrieval results obtained using the synthesized sketches. Numbers in red and blue indicate the best and second-best performance among compared models. The results are in top-1 and top-10 accuracy. }
\centering
\label{tab:recog}
\end{table}

\subsection{Sampling the Latent Space}
\begin{figure}[t]
\centering
\includegraphics[width=0.45\textwidth]{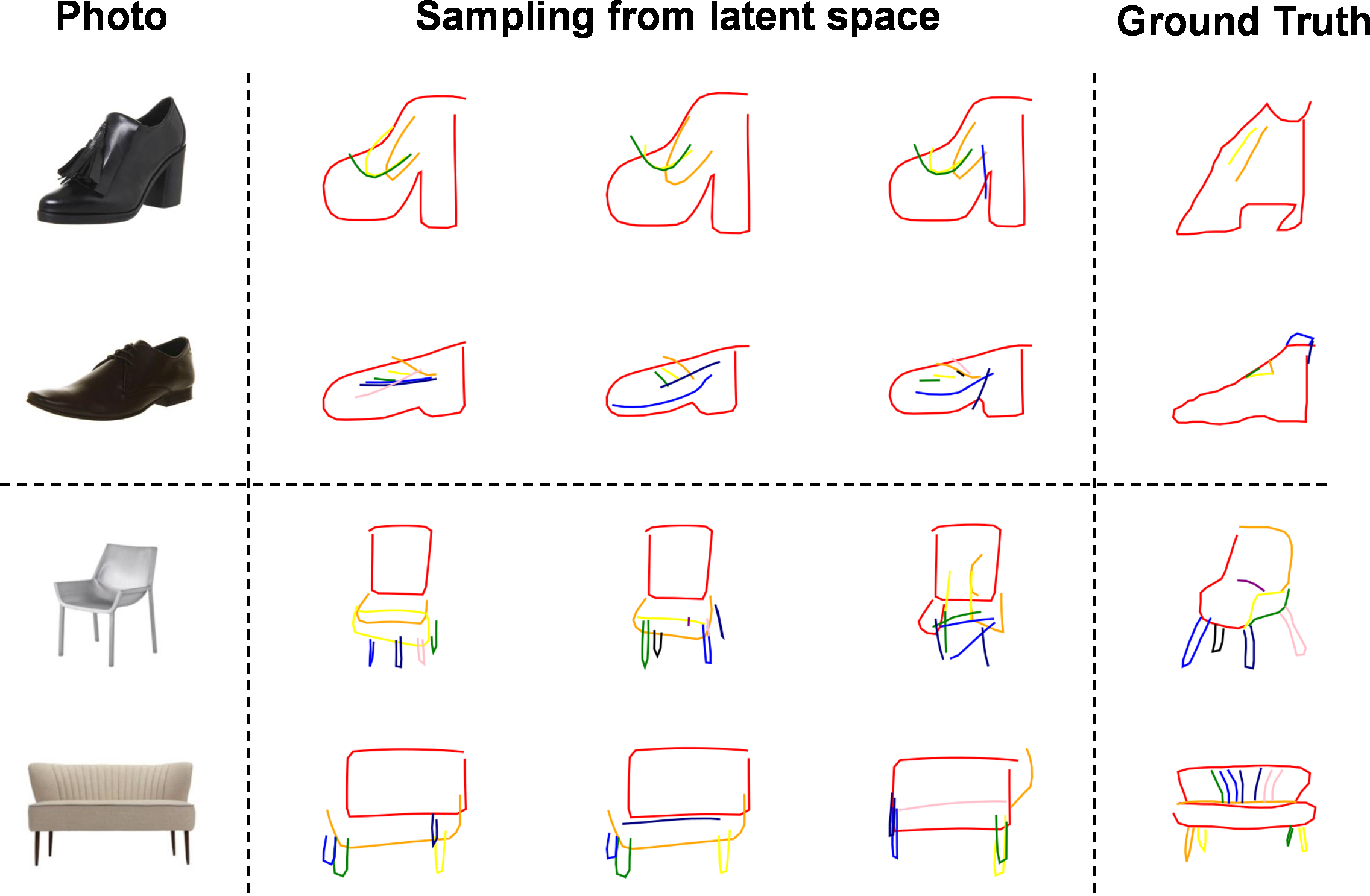}
\caption{Examples of different sketches synthesized for the same photo input by sampling the latent space with our full model.}
\label{fig:sampling}
\end{figure}

\textcolor{black}{With the help of the KL loss, we are able to exploit the embedding space from CNN encoder $E_p$ by effectively sampling from the latent vector $z$. It is thus intuitive that given one photo, our model can generate multiple sketches, as illustrated in Figure~\ref{fig:sampling}. We further observe that by re-sampling of the latent space, different synthesized sketches corresponding to the same reference photo can still keep instance-identifiable visual characteristics globally, but with differences at various local strokes/parts.}

\subsection{Data Augmentation for FG-SBIR}
\label{sub:augment}

In this experiment, we evaluate whether the synthesized sketches using our model can be used to form some additional photo-sketch pairs to train a better FG-SBIR model.  More concretely,  we collect 1800 photos from a different shopping website (Selfridge's), called ShoeSF, which have no overlap with the ShoeV2 photos. We then apply our model trained on ShoeV2 to  generate sketches for ShoeSF to form some additional photo-sketch pairs. They are then used to pretrain the FG-SBIR model in \cite{yu2016sketch} before fine-tuning on the ShoeV2 provided photo-sketch pairs.  Table~\ref{tab:fgsbir-syn} shows that using the synthesized data can boost the performance by 2.10\% acc.@1. 

\begin{table}[t]
\normalsize
\centering
\resizebox{0.9\columnwidth}{!}{
\begin{tabular}{l|c|c}
\hline
Dataset & acc.@1 & acc.@10  \\
\hline
Without  pretraining on synthetic data  & 30.33\% & 76.28\% \\
With  pretraining on synthetic data & 32.43\% & 77.48\% \\
\hline
\end{tabular}}
\protect\caption{Evaluation of the contribution of synthetic sketch pretraining on FG-SBIR.}
\centering
\label{tab:fgsbir-syn}
\end{table}

\section{Conclusion}

We  proposed the first deep stroke-level photo-to-sketch synthesis model that enables abstract stroke-level visual understanding of an object in a photo. To cope with the noisy supervision of photo-human sketch pairs, we proposed a novel supervised-unsupervised hybrid model with shortcut cycle consistency. We show that our model achieves  superior performance both qualitatively and quantitatively over a number of state-of-the-art alternatives. We also applied  our synthetic sketches as a mean of data augmentation for the FG-SBIR task, obtaining  promising results.


{\small
\bibliographystyle{ieee}
\bibliography{egbib}
}

\end{document}